\documentclass{sig-alternate}
\usepackage{algorithm}
\usepackage{algpseudocode}
\usepackage{amsmath}
\usepackage{graphics}
\usepackage{epsfig}

\begin{document}
%

\title{Latent Feature Based FM Model For Rating Prediction}
%
%
%
%
%

\numberofauthors{3} 
%
\author{
%
%
\alignauthor
Xudong Liu, Bin Zhang\\
	   \affaddr{ National Laboratory of Pattern Recognition}\\
       \affaddr{Institute of Automation}\\
       \affaddr{Chinese Academy of Sciences}\\
       \affaddr{Beijing, 100190, China}\\
       \email{\{lxdwinwin, sjlmustc\}@gmail.com}
\alignauthor
Ting Zhang\\
	   \affaddr{University of Science and Technology of China}\\
	   \affaddr{Hefei, 230026. China}\\
	  \email{ting.zhang.email@gmail.com}
\alignauthor
Chang Liu\\
       \affaddr{Alibaba Group}\\
       \affaddr{Beijing, 100022, China}\\
       \email{qingsong.lc@alibaba-inc.com}
}

\maketitle
\begin{abstract}
Rating Prediction is a basic problem in Recommender System, and one of the most widely used method is Factorization Machines(FM). However, traditional matrix factorization methods fail to utilize the benefit of implicit feedback, which has been proved to be important in Rating Prediction problem. In this work, we consider a specific situation, movie rating prediction, where we assume that a user's watching history has a big influence on his/her rating behavior on an item. We introduce two models, Latent Dirichlet Allocation(LDA) and word2vec, both of which perform state-of-the-art results in training latent features. Based on that,
we propose two feature based models. One is the Topic-based FM Model which provides the implicit feedback to the matrix factorization, the other is the Vector-based FM Model which exploits the order info of a user's watching history resulting in better performance. Empirical results on three datasets demonstrate that our method performs better than the baseline model and confirm that Vector-based FM Model usually works better as it contains the order info.
\end{abstract}





\section{Introduction}
The collaborative filtering problem has gained significant attention in machine learning field since the Netflix Prize. In this challenge, of the most widely used is the latent factor model which has proven to work well.
To state the problem more formally, we introduce several notations, that is, we have a set of users, $\mathcal{U} = \{U_1,U_2,\cdots,U_N\}$, a set of items, $\mathcal{I} = \{I_1,I_2,\cdots,I_M\}$, and the rating scores which can be viewed as a sparse matrix $\mathbf{R} \in \mathcal{R}^{N \times M}$, where the element $r_{ui}$ is the score rated by user $U_u$ on item $I_i$. 
The goal of the problem is to reasonably predict the missing elements in the sparse matrix.

Many methods have been designed to address that problem. Here, we mainly focus on matrix factorization \cite{Rendle:FM} as it performs the state-of-the-arts in dyadic prediction problem. However, matrix factorization \cite{Koren:MF} fails to utilize the benefit of the implicit feedback \cite{Koren:IF}, which plays an important role in recommender system. In order to provide implicit feedback to the matrix factorization, SVD++ \cite{Koren:SVD++} model is proposed but it takes much longer time and larger memory in the training process. Factorization Machines (FM) can be regarded as a classification or regression model combined with feature engineering. With different features, FM can mimic different factorization models like matrix factorization and specialized models such as SVD++. In this work, we propose two latent feature based FM models, both of which can get the implicit feedback of a user or an item. One is called Topic-based FM Model, and the other is Vector-based FM Model.

Topic model is a typical statistical model in natural language processing (NLP) area and also used in machine learning area. One of the most classic models is Latent Dirichlet Allocation (LDA)\cite{Blei:LDA}, which is a generative probabilistic model for collections of discrete data such as text corpora. It assumes that each document can be expressed by several topics, and each topic is generated by some words. As a result, we can express a document using a latent topic factor. Besides that, LDA can also be applied to a rating prediction problem. Consider a specific situation, where we want to predict how a user will rate a movie based on user's watching history which, to some degree, can indicate user's interest on this movie. Thinking of a user's watching history as a ``document'' and each movie as a ``word'' in this ``document'', we can see that a user's interest can be similarly obtained by several latent topics, which are generated by the ``words'' that belong to the ``document''. In other words, the user's interests in movies can be drawn from those latent topics. Therefore, we can use the latent topics as features to train the FM model, which we called Topic-based FM model.

The other FM model we proposed is Vector-based FM Model which is built on word2vec\cite{Mikolov:word2vecNIPS}.
It provides an efficient implementation of continuous bag-of-words(CBOW) and skip-gram architectures for computing vector representations of words\cite{Mikolov:word2vec}. Though it is a simple neural network model, it works quite well in practice.
The main goal of word2vec is to introduce techniques that can be used for learning high-quality word vectors from huge data sets with billions of words, and a big vocabulary with millions of words. Similar to LDA model, word2vec model can also train a latent vector from a vocabulary constructed from the training data. The difference from LDA is that this latent vector represents the word itself instead of the document. 
In our problem, a user is regarded as a ``document'', and the user's watching history can be viewed as a sequence of ``words''. Thus word2vec model can be used here to generate a latent vector for each item. Following the same fashion, we use those latent vectors as features to train the FM model resulting in Vector-based FM Model. 

The following of this paper is organized as follows. In Section 2, we provide more detailed description on Topic-based FM Model and Vector-based FM Model. Section 3 shows experimental evaluation and analysis of our method on three large scale collaborative datasets, which demonstrates that our method outperforms state-of-the-art latent factor approaches. Finally, we conclude in Section 4.

\section{Latent Feature Based FM Model in Rating Prediction}
In this section, we will introduce the two latent feature models mentioned above into rating prediction problem. These latent features may bring some implicit feedback or some latent characters of a user or an item. In the following part, we will explain how the latent features work in FM model.\\

\subsection{Topic-based FM Model}
Topic-based FM Model is similar to a previous work, $M^3F$ model \cite{Mackey:m3f} , which generates latent factors from user's history info and item's history info. However, in $M^3F$ model, the latent factors must be trained every once a time which is time-consuming. Our work, on the contrary, doesn't need to train the latent factors every time. We just need to generate the latent factors for the first time and update them when necessary, leading to a simpler algorithm. Next we will explain how topic model works in FM model and show the detailed algorithm.

First we introduce the $M^3F$ model to make the notation clear here and below. The $M^3F$ model takes three steps to obtain the parameters of user and item using Gibbs Sampling. It firstly samples the hyperparameters, then samples the topics and finally the user parameters and item parameters. For more details about the three steps, you may refer to the paper \cite{Mackey:m3f}. In general, the $M^3F$ model introduced two methods to predict the missing elements in the rating matrix. One is the $M^3F$-TIB model, the predicted score rated by user $U_u$ on item $I_i$ is obtained by the following formula,
\begin{align}
	 \hat{r}_{ui} (\vec \theta_u, \vec \theta_i) =\mathbf{p}_u \cdot \mathbf{q}_i + \sum_{k=1}^{K_U}\theta_{uk}w_{uk} + \sum_{l=1}^{K_I}\theta_{il}w_{il}.
\end{align}
$\mathbf{p}_u$ and $\mathbf{q}_i$ are the latent vectors for the user $U_u$ and item $I_i$ respectively. $\mathbf{p}_u\cdot \mathbf{q}_i$ represents the dot product between the two vectors.
$\vec \theta_u = [\theta_{u1}, \theta_{u2}, \cdots, \theta_{uK_U}]^T$ is the latent topic for the corresponding user and $\mathbf{w}_u = [w_{u1},w_{u2},\cdots, w_{uK_U}]^T$ is the weight vector for the latent topic.
Similarly, $\vec \theta_i$ and $\mathbf{w}_i$ are the latent topic for item $I_i$ and weight vector for that latent topic.
The other one is $M^3F$-TIF model whose prediction formula is given as follows,
\begin{align}
	 \hat{r}_{ui} (\vec \theta_u, \vec \theta_i) =\mathbf{p}_u\cdot \mathbf{q}_i + \sum_{k=1}^{K_U} \sum_{l=1}^{K_I}\theta_{uk}\theta_{il}\mathbf{e}_{uk}\cdot \mathbf{e}_{il},
\end{align}
where $\mathbf{e}_{uk}$ and $\mathbf{e}_{il}$ are the topic-indexed vectors for $\theta_{uk}$ and $\theta_{il}$ respectively. They provide the weight for the user-item-cross $\theta_{ui}\theta_{il}$ using the dot product.

In our formulation, we solve the problem by combining those two existing methods mentioned above. Firstly, we train the user's latent factor based on the user's history info. Secondly, we train the item's latent factor using the item's history info which tells those users who have watched this item.
After we get the user's topic and the item's topic, we define our prediction formula based on FM as follows,
\begin{align}
	\hat{r}_{ui}(\vec \theta_u, \vec \theta_i) = & \mu + b_u + b_i + \mathbf{p}_u\cdot \mathbf{q}_i  + \sum_{k=1}^{K_U}\theta_{uk}w_{uk} + \sum_{l=1}^{K_I}\theta_{il}w_{il} \nonumber \\
	& + 
	\sum_{k < l}^{K_U}\theta_{uk}\theta_{ul}\mathbf{e}_{uk}\cdot \mathbf{e}_{ul} +
	\sum_{k < l}^{K_I}\theta_{ik}\theta_{il}\mathbf{e}_{ik}\cdot \mathbf{e}_{il}\\
	&+\sum_{k=1}^{K_U} \sum_{l=1}^{K_I}\theta_{uk}\theta_{il}\mathbf{e}_{uk}\cdot \mathbf{e}_{il}.
\end{align}
$\mu$ is the global bias, and ${b_u}$, ${b_i}$ are the user's and item's bias, $\sum_{k<l}^{K_U}$ stands for $\sum_{k=1}^{K_U}\sum_{l = k+1}^{K_U}$. 

Compared with $M^3F$ model, our method is different in two ways: $1)$, besides the cross terms between user and item, we add user-user-cross terms and item-item-cross terms to make our formulation containing more info, thus having more accurate results; $2)$, our method can be divided into two steps, training the latent factors and putting them into FM model. Therefore, we don't need to sample from history data for each one, which takes longer time. More details about our method is shown in Algorithm 1.

\begin{algorithm}[htb]
	\caption{ Topic based FM model}
	\begin{algorithmic}[1]	
		\For {all users $U_u \in \mathcal{U}$}
			\For { all items $I_i$ belong to $U_u$'s history}
				\State sample user topic index $ \tilde{\theta}_{uk} \thicksim P(\theta_{uk}|\vec \theta_{u\rightharpoondown k}, \mathcal{I})$
			\EndFor
		\EndFor 			
		\For {all items $I_i \in \mathcal{I}$}
			\For { all users $U_u$ belong to $I_i$'s history}
				\State sample item topic index $ \tilde{\theta}_{il} \thicksim P(\theta_{il}|\vec \theta_{i\rightharpoondown l}, \mathcal{U})$
			\EndFor
		\EndFor
		\State train FM model with $\vec \theta_u$ and $\vec \theta_i$
	\end{algorithmic}
	\label{alg:topicBasedModel}
\end{algorithm}

\begin{figure}
	\centering
	\epsfig{file=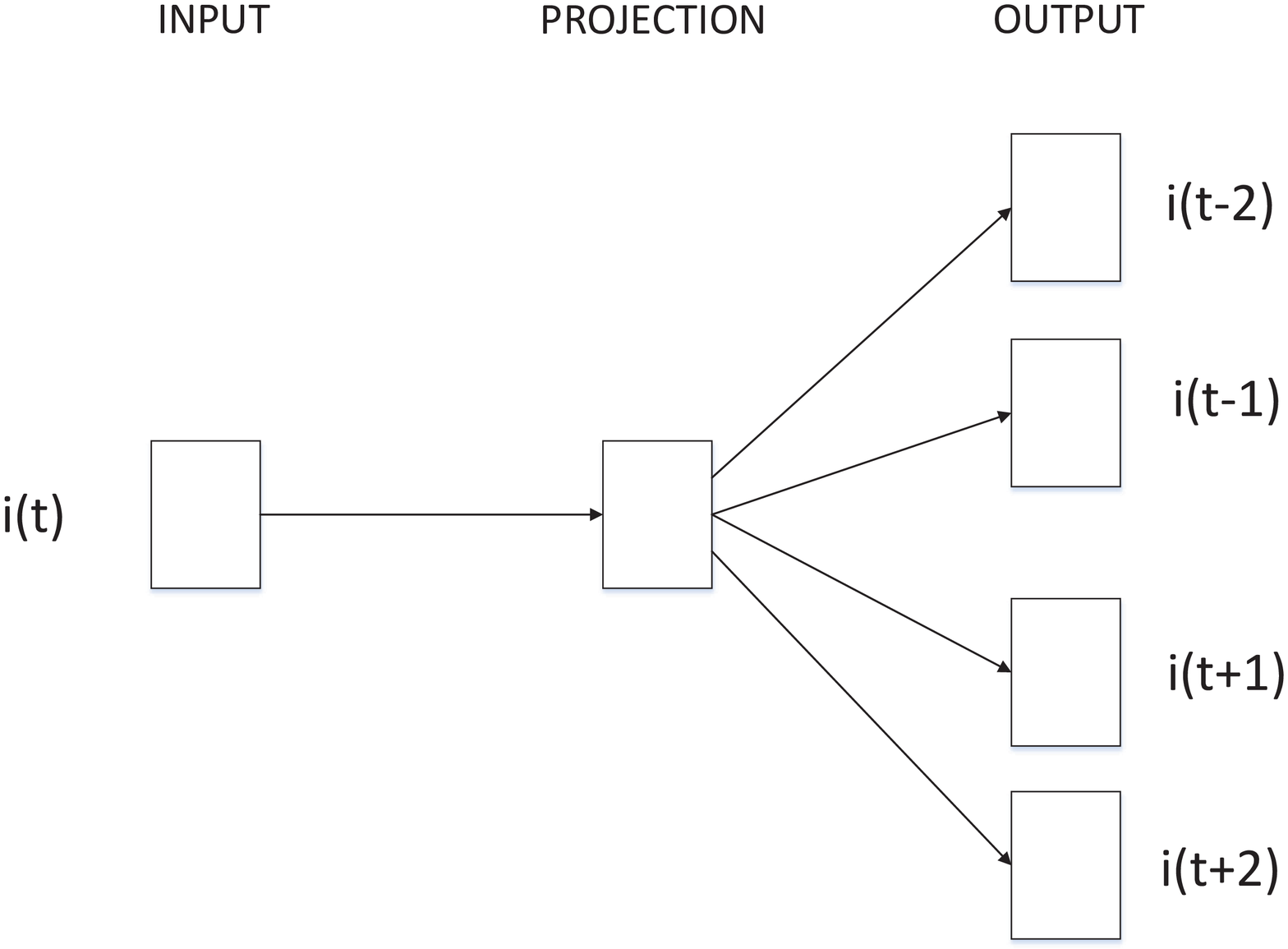,height=1.7in, width=3in}
	\caption{Framework of Skip Gram Model. i(item) represents an item in a user's watching history and items are ordered by time.}
\end{figure}

\begin{table*}
	\centering
	\caption{RMSE scores for Topic-based FM model on Baidu Data, 10M MovieLens and Netflix Prize}
	\begin{tabular}{c|c|c|c} \hline
		
		&Baidu Data&10M MovieLens&Netflix Prize \\ \hline
		method&iter=100 iter=200 iter=300&iter=100 iter=200 iter=300&iter=100 iter=200 iter=300 \\ \hline
		baseline&0.629178 0.628903 0.629111&0.788312 0.787922 0.787916&0.882469 0.879894 0.879355 \\
		topic\_8&0.626627 0.626536 0.626536&0.787557 0.787020 0.786987&0.871300 0.869120 0.868806\\
		topic\_20&\textbf{0.625879 0.625958 0.625944}&\textbf{0.787204 0.786677 0.786583}&\textbf{0.868443 0.866720 0.866411} \\
		\hline\end{tabular}
\end{table*}

\begin{figure*}[t]
	\centering
	(a)~~\includegraphics[width=.45\linewidth, clip]{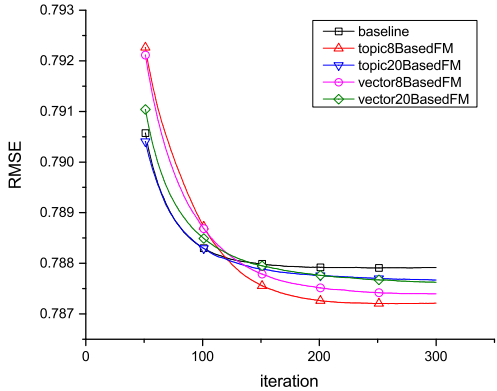}~~~~
	(b)~~\includegraphics[width=.45\linewidth, clip]{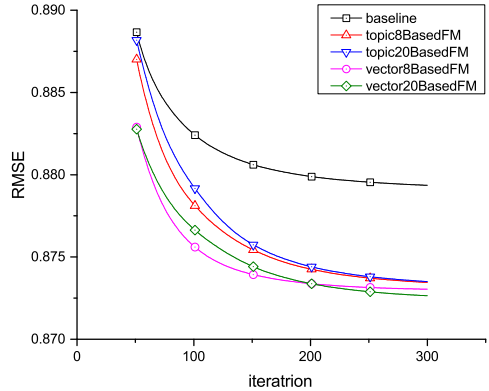}
	\caption{RMSE scores for Vector-based FM model on (a) 10M MovieLens dataset, (b) Netflix Prize dataset, iteration from 50 to 300 }
	\label{fig:epsilon0}
\end{figure*}

\subsection{Vector-based FM Model}
In the Topic-based FM Model, it views the items that belong to the same user as a set, so it fails to exploit the order of one user's watching history. In fact, one can easily observe that the items next to each other share some similarities to a certain extent since they indicate a user's interest in a short period of time. 
Driven by this observation, we apply the word2vec model which utilize the words order in a document to a rating prediction problem where user's watching history can be regarded as a ``document'' and the item in his/her watching history can be viewed as the ``word'' in that ``document''. In this way, The latent vector of the item  trained in this model represents users' interests to some degree.

So in order to get a better result, our approach, named Vector-based FM Model, takes advantage of that latent vector which is obtained by using Skip Gram method. The framework of Skip Gram is shown in Figure 1. We maximize the average log probability $\frac 1 T \sum_{t=1}^T\sum_{-c\le j \le c, j\neq 0}\log p(i_{t+j}|i_t)$ to get the item latent vector.

After that, we use the item latent vector as features in the FM model, so our prediction formula is,
\begin{align}
\hat{r}_{ui}(\mathbf{v}_i) = & \mu + b_u + b_i + \mathbf{p}_u\cdot \mathbf{q}_i  +  \sum_{l=1}^{K_I}v_{il}w_{il} \nonumber \\
& + \sum_{k < l}^{K_I}v_{ik}v_{il}\mathbf{e}_{ik}\cdot \mathbf{e}_{il},
\end{align}
where $\mathbf{v}_i = [v_{i1}, v_{i2}, \cdots,v_{iK_I}]^T$ is the latent item vector for the corresponding item.

In this formulation, we didn't train a user's latent vector as it can be seen that the ordered users who watched the same item don't have any strong relevance. In addition to train the order info of the watching history, word2vec is able to handle huge data sets with a big vocabulary. So our approach, Vector-based FM Model, can be well applied to large scale datasets.

\section{Experiment}
We evaluate our models on several large scale movie rating collaborative filtering datasets including the Baidu Movie Recommendation Algorithm Contest Data (Baidu Data for short)\footnote{http://openresearch.baidu.com/topic/40.jspx}, the Netflix Prize dataset\footnote{http://www.netflixprize.com/} and the 10M MovieLens Datasets\footnote{http://www.grouplens.org/}. The Baidu Data contains 1.26 million ratings on the scale from 1 to 5 distributed across 9,722 users and 7,889 movies. The 10M MovieLens Datasets contains 10 million ratings on the scale from 0.5 to 5 with half-star increments with 71,567 users and 10,681 movies. The Netflix Prize dataset has 480,189 users and 17,770 items and 100 million ratings in \{1,...,5\}. In our experiments, we use all the three datasets to evaluate our Topic-based FM Model. And we use only 10M MovieLens dataset and Netflix Prize dataset to compare the performance of the baseline model, topic based model and vector based model as Baidu Data doesn't contain time info. The baseline model is obtained only using user\_id and item\_id to train the FM model and the parameters for FM are fixed for each experiment. Finally we use RMSE to measure the model's performance.

\subsection{Experiments on Topic based FM}
We first introduce the experiments on Topic-based FM Model, for which we use a off-the-shelf software implementation, LibFM tool\cite{Rendle:libfm}. The default parameter setting is used in LibFM, where the dimension of the latent factor is 8 and the learning method is mcmc \cite{Heinrich:gibbslda}.

For LDA model, we also use an open source tool Gibbs LDA++\footnote{http://sourceforge.net/projects/gibbslda/} to generate the latent factor. We consider two experimental settings where the dimension of the latent factor is 8 and 20 separately, that is $K_U = K_I =8$ and $K_U = K_I =20$. For other parameters, we set alpha$=0.5$, beta$=0.1$ and iterations$=300$. 

Table 1 reports the performance of Topic-based FM Model, which shows a significant improvement on RMSE. Among the three methods, Topic-based FM Model with 20 latent factors performs best on all the three datasets, especially on Baidu Data and Netflix Prize datasets. As expected, when the dimension of latent factors increases, the model will have more expressive ability on user or item, thus the performance in terms of RMSE is better, which is also demonstrated by our experiment results.

\subsection{Experiments on Vector based FM}
As mentioned above, the user's latent vector is not considered in Vector-based FM Model, and only the item's latent vector is trained. So for a fair comparison, only the item's topic is used in Topic-based FM Model. Then the results on both methods are compared with the baseline model. In this experimental setting, we use a publicly available implementation of word2vec\footnote{https://code.google.com/p/word2vec/}. The dimension of latent vector is set to $K_U = K_I =8$ and $K_U = K_I = 20$. The time window is set to 3, which means the prior 3 items and the posterior 3 items are considered in the training process. And we use the Skip Gram method to train.


Figure 2 shows the convergence curve on 10M MovieLens dataset and Netflix Prize dataset. We can see that RMSE of both of our proposed approaches is lower than the baseline. On MovieLens dataset, our two latent feature based FM models converge slower than baseline model, but a lower RMSE level is achieved than baseline. In addition, comparable results are obtained by Topic based and Vector based FM model.
On Netflix Prize dataset, we can see that both of our methods not only perform better than baseline, but also converge more quickly. The conclusion is that Vector-based FM model performs better since it exploits the order of watching history is validated by our experiments.


\section{Conclusion}
In this work, we introduce topic based latent features and vector based latent features into traditional FM model resulting in two feature based models. The Topic-based FM Model provides the implicit feedback within less training time since we only need to update when necessary. The Vector-based FM Model exploits the order info of a user's watching history resulting in better performance. Empirical results on three datasets demonstrate that our method performs better than the baseline model and confirm that Vector based FM model usually works better as it contains the order info. For the future work, to improve the performance, we may adjust parameters and generate latent features which better express the users or items.

\bibliographystyle{abbrv}
\bibliography{reference}
\end{document}